\pdfoutput=1
\documentclass[11pt]{article}

\usepackage[preprint]{acl}

\usepackage{times}
\usepackage{latexsym}

\usepackage[T1]{fontenc}
\usepackage[utf8]{inputenc}

\usepackage{microtype}

\usepackage{inconsolata}

\usepackage{graphicx}
\usepackage{subcaption}
\usepackage{enumitem}
\setlist[itemize]{noitemsep,nolistsep}
\usepackage{booktabs}
\usepackage{multirow}
\usepackage{makecell}
\usepackage{amsmath}
\usepackage{bbding}
\usepackage{algorithm}
\usepackage{algpseudocode}
\usepackage{xcolor}
\usepackage{microtype}
\usepackage{colortbl}
\usepackage{pifont}
\usepackage{url}
\usepackage{subcaption}
\usepackage{adjustbox}
\usepackage{amsfonts}
\usepackage{afterpage}
\usepackage{ulem}
\usepackage{CJKutf8}
\usepackage{pgf-pie}  % 画扇形图的宏包

\usepackage[listings,skins,breakable]{tcolorbox}
\usepackage{multicol}
\usepackage{verbatim}
\definecolor{CandyPink}{RGB}{255, 182, 193}
\definecolor{SoftBlue}{RGB}{173, 216, 230}
\definecolor{MintGreen}{RGB}{152, 255, 152}
\definecolor{PeachOrange}{RGB}{255, 218, 185}
\definecolor{LavenderPurple}{RGB}{230, 190, 255}
\definecolor{softpink}{RGB}{240, 250, 255}
\definecolor{uclablue}{RGB}{159, 195, 224}
\definecolor{darkuclablue}{rgb}{0.0, 0.35, 0.55} 

\definecolor{lightred}{RGB}{200, 160, 160}
\definecolor{lightred2}{RGB}{230, 100, 100}
\definecolor{lightorange}{RGB}{255, 165, 79}
\definecolor{softdarkblue}{RGB}{70, 130, 180}
\definecolor{uclagold}{RGB}{255, 240, 180}

\definecolor{mycell}{gray}{.95}
\definecolor{mycelltwo}{RGB}{255, 182, 193}
\definecolor{ForestGreen}{rgb}{0.13, 0.55, 0.13}
\definecolor{DarkCoral}{rgb}{0.72, 0.33, 0.31}
\definecolor{lightgray}{RGB}{230, 230, 230}
\definecolor{babygreen}{rgb}{0.85, 0.97, 0.85}

\newcommand{\yes}{\color{green!60!black}\ding{51}}
\newcommand{\no}{\color{red!60!black}\ding{55}}

\newcommand{\addMethod}[1]{~~\textsl{+ #1}}
\newtcolorbox{titleEnv}{
colframe=black!80,
colback=gray!10,
fonttitle=\bfseries,
coltitle=black,
left=3pt,
right=3pt,
top=3pt,
bottom=3pt,
boxrule=0.4mm,
arc=3mm
}
\definecolor{codegreen}{rgb}{0,0.6,0}
\definecolor{codegray}{rgb}{0.5,0.5,0.5}
\definecolor{codepink}{RGB}{252, 142, 172}
\definecolor{codepurple}{rgb}{0.58,0,0.82}
\definecolor{backcolour}{RGB}{245,245,245}
\definecolor{royalblue(web)}{rgb}{0.25, 0.41, 0.88}
\definecolor{whitesmoke}{rgb}{0.96, 0.96, 0.96}
\lstdefinestyle{mystyle}{
    language=Python,
    commentstyle=\color{codegreen},
    keywordstyle=\color{magenta},
    numberstyle=\tiny\color{codegray},
    stringstyle=\color{codepurple},
    basicstyle=\ttfamily \lst@ifdisplaystyle\tiny\fi,
    breakatwhitespace=false,         
    breaklines=true,                 
    captionpos=b,                    
    keepspaces=true,                 
    numbers=left,                    
    numbersep=5pt,                  
    xleftmargin=12pt,
    showspaces=false,                
    showstringspaces=false,
    showtabs=false,                  
    tabsize=2,
    moredelim=[is][\bfseries]{<highlight>}{</highlight>}, %
    postbreak=\raisebox{0ex}[0ex][0ex]{\ensuremath{\color{black}\lst@ifdisplaystyle\hookrightarrow\fi\space}} %
}

\newtcolorbox{findings}[1][]{
	float,
  	title=#1,
	colframe=darkuclablue,
        top=1pt,           % 控制顶部空白
        bottom=1pt,        % 控制底部空白
        left=0pt,          % 控制左边空白
        right=0pt,          % 控制右边空白
        before skip=0.65em, after skip=0.75em,
}

\newtcolorbox{promptbox}[2][Prompt]{
colback=black!5!white,
arc=5pt, 
boxrule=0.5pt,
fonttitle=\bfseries,
title=#1, 
before upper={\small}, fontupper=\fontfamily{ptm}\selectfont,
colframe=#2, % 使用传递的参数来设定 colframe
}

\newtcolorbox{insights}[1][]{
	float,
  	title=#1,
        top=1pt,           % 控制顶部空白
        bottom=1pt,        % 控制底部空白
        left=0pt,          % 控制左边空白
        right=0pt,          % 控制右边空白
        before skip=0.65em, after skip=0.75em,
}
\title{
Benchmarking Temporal Reasoning and Alignment Across Chinese Dynasties
}

\author{
    Zhenglin Wang\thanks{~~Equal Contribution. The work was partially done during Jialong's internship at Alibaba Group.}$^{\heartsuit}$, \hspace{0.5mm}
    Jialong Wu$^{*}$$^{\heartsuit}$${^\diamondsuit}$,\hspace{0.5mm}
    Pengfei Li$^{\heartsuit}$,\hspace{0.5mm}
    Yong Jiang$^{\diamondsuit}$,\hspace{0.5mm}
    Deyu Zhou$^{\heartsuit}$\thanks{~~Corresponding Author.}\hspace{0.5mm}
    \\
    $^{\heartsuit}$\hspace{0.5mm}School of Computer Science and Engineering, Key Laboratory of Computer Network\\
    and Information Integration, Ministry of Education, Southeast University, China \\
    $^{\diamondsuit}$\hspace{0.5mm} Tongyi Lab, Alibaba Group\\
    \texttt{\{zhenglin, jialongwu, d.zhou\}@seu.edu.cn}
} 

\begin{document}
\maketitle
\begin{abstract}
% Large Language Models (LLMs)...
Temporal reasoning is fundamental to human cognition and is crucial for various real-world applications. 
While recent advances in Large Language Models have demonstrated promising capabilities in temporal reasoning, existing benchmarks primarily rely on rule-based construction, lack contextual depth, and involve a limited range of temporal entities. 
To address these limitations, we introduce \textbf{C}hinese \textbf{T}i\textbf{m}e Reasoning (CTM), a benchmark designed to evaluate LLMs on temporal reasoning within the extensive scope of Chinese dynastic chronology.
CTM emphasizes cross-entity relationships, pairwise temporal alignment, and contextualized and culturally-grounded reasoning, providing a comprehensive evaluation. 
Extensive experimental results reveal the challenges posed by CTM and highlight potential avenues for improvement.\footnote{Code and dataset are available at~\url{https://github.com/Linking-ai/ctm_bench}}

% Our dataset and the source code are available at~\url{https://github.com/Linking-ai/Crosstempbench}
\end{abstract}

\begin{CJK*}{UTF8}{gkai}
\section{Introduction}
\begin{quote}
``究天人之际，通古今之变。''\\
\hspace*{\fill}--- 司马迁《史记·报任安书》
\end{quote}
Understanding time is fundamental to human cognition and plays a pivotal role in shaping our perception and interaction with the world~\cite{islakoglu2025chronosense}.
Recently, Large Language Models (LLMs)have shown promising abilities in temporal reasoning~\citep{chu-etal-2024-timebench,su-etal-2024-living}.
\begin{figure}[t]
    \centering
    \includegraphics[width=0.45\textwidth]{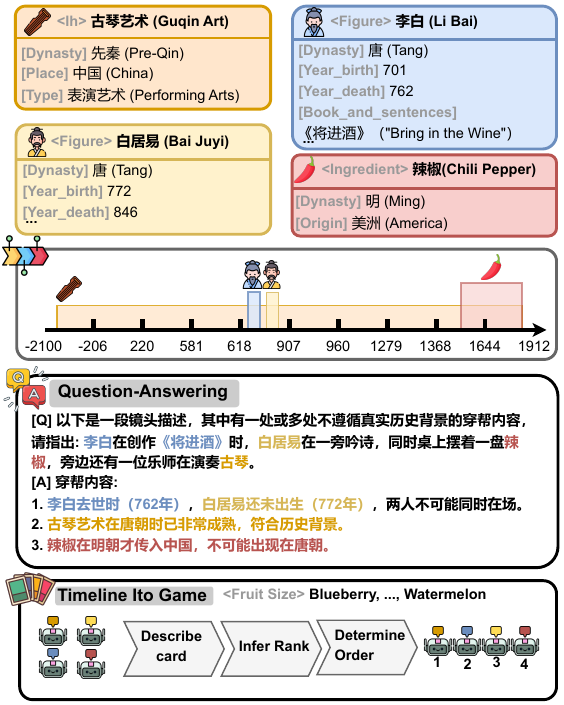}
    \caption{A QA pair from a script error correction task and an instance of the Timeline Ito Game with a ``\textit{fruit size}'' theme from CTM.
    \protect\footnotemark
    }
    \label{fig:intro}
\end{figure}
\footnotetext{The English translation is presented in App.~\ref{app:tran_qa}.
}
Previous benchmarks, which rely on rule-based constructed methods, lack contextualization and involve a limited number of entities in temporal relation evaluation.
The core principle in assessing temporal reasoning lies in evaluating whether the model has a clear understanding of the event time within a temporal coordinate system. 
Compared to other temporal coordinate systems, the Chinese dynastic chronology spans a significantly longer historical scope and encompasses a broader range of culturally-grounded and historical knowledge~\citep{sun-etal-2024-benchmarking-chinese, li-etal-2024-foodieqa, yuan2024cultural, lu2024benchmarking}.
It serves as a well-suitable background for temporal reasoning, as real-world applications can be found in various media, including films, short dramas, and novel writing, all of which rely on it.

\begin{table*}[t]
\small
\centering
\caption{\textbf{Comparison} between CTM and other benchmarks.
Detailed discussion is presented in Appendix~\ref{app:rw}.}
\resizebox{1.8\columnwidth}{!}{%
\begin{tabular}{@{}c|c|c|c|c|c|c@{}}
\toprule
& \textbf{Language}  & \textbf{Construction} & \textbf{Time Scope} & \textbf{Contextualization} & \textbf{Temporal Alignment}
& \textbf{Complex Aspects}\\
\toprule
\textsc{TimeQA}~\shortcite{chen2021a} & En & Rule-based & 1367--2018 & \no & \no & \no\\
\textsc{TempLAMA}~\shortcite{dhingra-etal-2022-time} & En & Rule-based & 2010--2020 & \no & \no & \no\\
\textsc{TempReason}~\shortcite{tan-etal-2023-towards} & En & Rule-based & 634--2023 & \no & \no & \no\\
\textsc{SituatedGen}~\shortcite{zhang2023situatedgen} & En & LLM-based & - & \yes & \no & \yes \\
\textsc{CoTemp\-QA}~\shortcite{su-etal-2024-living}  & En & Rule-based & - & \no & \no & \no\\
\textsc{TimeBench}~\shortcite{chu-etal-2024-timebench}  & En & - & - & \yes & \no & \yes\\
\textsc{TRAM}~\shortcite{wang-zhao-2024-tram} & En & Rule-based & - & \yes & \no & \yes\\
\textsc{ChronoSense}~\shortcite{islakoglu2025chronosense}  & En & Rule-based & - & \no & \no & \no\\
\textbf{CTM}  & Zh & LLM-based & -2100--1912 & \yes & \yes & \yes\\
\bottomrule 
\end{tabular}
}
\label{table:comparsion}
\end{table*}

Therefore, we introduce \textbf{C}hinese \textbf{T}i\textbf{m}e Reasoning (\textbf{CTM}) benchmark in this study.
The comparison between CTM and other benchmarks is shown in Table~\ref{table:comparsion}.
\textbf{CTM} focuses on contextualization, cross-entity relationships, and pair-wise temporal alignment capability.
As shown in Figure~\ref{fig:intro}, answering this question requires a clear temporal understanding of four entities, ``李白'' (701 to 762), ``白居易'' (772 to 826), ``古琴'' (Since ``Pre-Qin''), and ``辣椒'' (Since ``Ming'').
In addition, we develop the Timeline Ito Game to evaluate the LLM's ability to align entities across temporal and other dimensions, requiring pairwise order perception of different entities.
The CTM benchmark is built upon a curated and authoritative Chinese cultural entity repository, which encompasses over 4,700 entities, spanning from figures, places, allusions, ingredients, and intangible cultural heritage.

We evaluate the performance of the CTM benchmark using various mainstream LLMs, including both closed-source and open-sourced from diverse perspectives.
We conduct experiments under both zero-shot and chain-of-thought (CoT) settings~\cite{wei2022chain}, respectively.
Further analysis shows the challenge of CTM and 
provides empirical insights into enhancing LLMs' temporal reasoning abilities and alignment across Chinese dynasties.

The contributions of this work are as follows: 1). We construct an interesting and challenging benchmark CTM, comprising \textbf{8,750} QA pairs and \textbf{60} instances of Timeline Ito Games.
2). We conduct extensive empirical experiments with various LLMs, which show that the proposed tasks are challenging.
3). Analysis of the experiments further reveals potential directions for addressing temporal reasoning tasks.
\section{CTM Dataset}

\subsection{Task Definition}
\begin{table*}[t]
\centering
\caption{ 
\textbf{Main results on QA tasks} within CTM benchmark.
The best results among all backbones are \textbf{bolded}, and the second-best results are \underline{underlined}.
}
\small
\setlength\tabcolsep{2pt}
\resizebox{0.90\textwidth}{!}{%

\begin{tabular}{l|ccccc|ccccccc|c}
\toprule

\multicolumn{1}{c|}{\multirow{3}{*}{\textbf{Method}}} & \multicolumn{5}{c|}{\textbf{Cross Temp Count}} & \multicolumn{7}{c|}{\textbf{Question Type}} & \multicolumn{1}{c}{\multirow{3}{*}{\textbf{Avg.}}} \\
\cmidrule{2-13}
 & $=1$ (EDD) & $=2$ & $=3$ & $\geq 4$ & $\geq 4_{L}$ (LSEC) & PJ & TOU & RR & SEC & EEU & TIC & TES\\
\midrule
\multicolumn{14}{c}{\cellcolor{uclablue} \textbf{\textit{Closed-Sourced LLMs}}} \\
\midrule
GPT-4o & 56.52 & 51.12 & 44.76 & 26.10 & 53.60
& 58.64 & 38.42 & 57.26 & 36.15 & 40.58 & 15.36 & 59.31 & 48.08\\
\addMethod{CoT} & \underline{67.40}\tiny\textcolor{red}{+10.88} & \underline{58.08}\tiny\textcolor{red}{+6.96} & 49.24\tiny\textcolor{red}{+4.48} & 29.60\tiny\textcolor{red}{+3.50} & 31.60\tiny\textcolor[HTML]{206546}{-22.0}
& \underline{64.10}\tiny\textcolor{red}{+5.46} & \underline{44.71}\tiny\textcolor{red}{+6.29} & \underline{59.62}\tiny\textcolor{red}{+2.36} & \underline{47.09}\tiny\textcolor{red}{+10.94} & 44.06\tiny\textcolor{red}{+3.48} & \underline{17.70}\tiny\textcolor{red}{+2.34} & \underline{61.68}\tiny\textcolor{red}{+2.37}
& \underline{54.21}\tiny\textcolor{red}{+6.13}\\

\arrayrulecolor{black!20}\midrule
Qwen-max & 60.48 & 53.12 & \underline{50.54} & 30.80 & \underline{62.00}
& \textbf{64.39} & 42.55 & 59.10 & 40.71 & \underline{46.38} & \textbf{20.87} & 60.22 & 52.27\\
\addMethod{CoT} & \textbf{69.56}\tiny\textcolor{red}{+9.08} & \textbf{59.32}\tiny\textcolor{red}{+6.20} & \textbf{54.48}\tiny\textcolor{red}{+3.94} & \underline{31.90}\tiny\textcolor{red}{+1.10} & 39.60\tiny\textcolor[HTML]{206546}{-22.40}
& 63.29\tiny\textcolor{red}{-1.10} & \textbf{48.58}\tiny\textcolor{red}{+6.03} & \textbf{63.75}\tiny\textcolor{red}{+4.65} & \textbf{55.77}\tiny\textcolor{red}{+15.06} & \textbf{53.91}\tiny\textcolor{red}{+7.53} & 15.19\tiny\textcolor{red}{-5.68} & \textbf{63.14}\tiny\textcolor{red}{+2.92} & \textbf{57.24}\tiny\textcolor{red}{+4.97} \\

\arrayrulecolor{black!20}\midrule
\rowcolor{mycell}
o1-preview & 52.80 & 46.56 & 49.64 & \textbf{32.70} & \textbf{67.20}
& 58.28 & 44.28 & 53.01 & 43.16 & 40.87 & 11.02 & 56.02 & 48.24\\

\arrayrulecolor{black}\midrule
\multicolumn{14}{c}{\cellcolor{uclagold} \textbf{\textit{Open-Sourced LLMs}}} \\
\midrule
LLaMA3.1$_{\text{8b}}$ & 33.04 & 16.86 & 15.60 & 9.10 & 10.80
& 19.66 & 12.95 & 18.65 & 7.37 & 0.87 & 2.01 & 37.04 & 20.14\\
\addMethod{CoT} & 35.05\tiny\textcolor{red}{+2.01} & 26.44\tiny\textcolor{red}{+9.58} & 19.96\tiny\textcolor{red}{+4.36} & 10.70\tiny\textcolor{red}{+1.60} & 12.40\tiny\textcolor{red}{+1.60}
& 26.48\tiny\textcolor{red}{+6.82} & 19.55\tiny\textcolor{red}{+6.60} & 23.20\tiny\textcolor{red}{+4.55} & 20.02\tiny\textcolor{red}{+12.65} & 15.70\tiny\textcolor{red}{+14.83} & 5.51\tiny\textcolor{red}{+3.50} & 34.37\tiny\textcolor[HTML]{206546}{-2.67} & 24.91\tiny\textcolor{red}{+4.77} \\

\arrayrulecolor{black!20}\midrule
ChatGLM3$_{\text{6b}}$ & 38.40 & 21.60 & 16.04 & 5.80 & 4.80
& 21.40 & 12.28 & 22.67 & 12.25 & 12.75 & 1.84 & 35.58
& 22.52\\
\addMethod{CoT} & 37.24\tiny\textcolor[HTML]{206546}{-1.16} & 22.72\tiny\textcolor{red}{+1.12} & 15.28\tiny\textcolor[HTML]{206546}{-0.76} & 8.20\tiny\textcolor{red}{+2.40} & 4.00\tiny\textcolor[HTML]{206546}{-0.80}
& 20.32\tiny\textcolor[HTML]{206546}{-1.08} & 15.92\tiny\textcolor{red}{+3.64} & 20.12\tiny\textcolor[HTML]{206546}{-2.55} & 14.98\tiny\textcolor{red}{+2.73} & 16.52\tiny\textcolor{red}{+3.77} & 3.01\tiny\textcolor{red}{+1.17} & 29.74\tiny\textcolor[HTML]{206546}{-5.84} 
& 22.61\tiny\textcolor{red}{+0.09} \\

\arrayrulecolor{black!20}\midrule
InternLM2.5$_{\text{7b}}$ & 60.64 & 47.32 & 39.36 & 21.60 & 42.00
& 51.39 & 30.16 & 48.64 & 45.78 & 42.61 & 11.19 & 50.18 
& 45.75\\
\addMethod{CoT} & 61.44\tiny\textcolor{red}{+0.80} & 51.40\tiny\textcolor{red}{+4.08} & 39.36\tiny\textcolor{red}{+0.00} & 20.20\tiny\textcolor[HTML]{206546}{-1.40} & 38.00\tiny\textcolor[HTML]{206546}{-4.00}
& 51.70\tiny\textcolor{red}{+0.31} & 31.45\tiny\textcolor{red}{+1.29} & 49.47\tiny\textcolor{red}{+0.83} & \underline{52.86}\tiny\textcolor{red}{+7.08} & 44.19\tiny\textcolor{red}{+1.58} & 11.52\tiny\textcolor{red}{+0.33} & 48.54\tiny\textcolor[HTML]{206546}{-1.64}
& 46.90\tiny\textcolor{red}{+1.15} \\

\arrayrulecolor{black!20}\midrule
Qwen2.5$_{\text{7b}}$ & 51.80 & 39.88 & 35.96 & 12.40 & 30.00
& 46.28 & 26.38 & 46.28 & 24.14 & 36.23 & 7.35 & 52.01 
& 38.76\\
\addMethod{CoT} & 59.96\tiny\textcolor{red}{+8.16} & 47.60\tiny\textcolor{red}{+7.72} & 36.64\tiny\textcolor{red}{+0.68} & 18.30\tiny\textcolor{red}{+5.90} & 30.80\tiny\textcolor{red}{+0.80}
& 52.46\tiny\textcolor{red}{+6.18} & 29.95\tiny\textcolor{red}{+3.57} & 52.18\tiny\textcolor{red}{+5.90} & 34.13\tiny\textcolor{red}{+9.99} & 40.58\tiny\textcolor{red}{+4.35} & 8.18\tiny\textcolor{red}{+0.83} & 49.64\tiny\textcolor{red}{-2.37} & 44.22\tiny\textcolor{red}{+5.46} \\

\arrayrulecolor{black!20}\midrule
Qwen2.5$_{\text{14b}}$ & 54.36 & 51.16 & 42.56 & 23.80 & 42.00 
& 57.44 & 36.86 & 51.83 & 36.90 & 39.07 & 18.26 & 58.58 
& 46.32\\
\addMethod{CoT} & 57.92\tiny\textcolor{red}{+3.56} & 45.44\tiny\textcolor[HTML]{206546}{-5.72} & 41.24\tiny\textcolor[HTML]{206546}{-1.32} & 22.50\tiny\textcolor[HTML]{206546}{-1.30} & 30.80\tiny\textcolor[HTML]{206546}{-11.20} & 
52.73\tiny\textcolor[HTML]{206546}{-4.71} & 34.36\tiny\textcolor[HTML]{206546}{-2.50} & 46.52\tiny\textcolor[HTML]{206546}{-5.31} & 42.57\tiny\textcolor{red}{+5.67} & 36.81\tiny\textcolor[HTML]{206546}{-2.26} & 10.02\tiny\textcolor[HTML]{206546}{-8.24} & 51.82\tiny\textcolor[HTML]{206546}{-6.76} & 
44.89\tiny\textcolor[HTML]{206546}{-1.43} \\

\arrayrulecolor{black!20}\midrule
Qwen2.5$_{\text{32b}}$ & 56.28 & 52.78 & 46.24 & 26.90 & 46.40
& 60.66 & 38.54 & 56.79 & 39.12 & 43.77 & 20.10 & 60.04 & 48.83\\
\addMethod{CoT} & 60.80\tiny\textcolor{red}{+4.52} & 49.32\tiny\textcolor[HTML]{206546}{-3.46} & 45.32\tiny\textcolor[HTML]{206546}{-0.92} & 24.80\tiny\textcolor[HTML]{206546}{-2.10} & 31.20\tiny\textcolor[HTML]{206546}{-15.20} & 
50.67\tiny\textcolor[HTML]{206546}{-9.99} & 40.65\tiny\textcolor{red}{+2.11} & 51.12\tiny\textcolor[HTML]{206546}{-5.67} & 43.40\tiny\textcolor{red}{+4.28} & 40.29\tiny\textcolor[HTML]{206546}{-3.48} & 17.03\tiny\textcolor[HTML]{206546}{-3.07} & 57.12\tiny\textcolor[HTML]{206546}{-2.92} 
& 48.14\tiny\textcolor[HTML]{206546}{-0.69} \\

\arrayrulecolor{black!20}\midrule
Qwen2.5$_{\text{72b}}$ & 58.20 & 48.76 & 46.84 & 31.30 & \underline{60.80} 
& 61.38 & 40.77 & 54.31 & 36.62 & 42.03 & 11.52 & \underline{62.23} & 49.30\\
\addMethod{CoT} & \underline{69.00}\tiny\textcolor{red}{+10.80} & \underline{57.24}\tiny\textcolor{red}{+8.48} & \underline{49.88}\tiny\textcolor{red}{+3.04} & \underline{32.50}\tiny\textcolor{red}{+1.20} & 46.00\tiny\textcolor[HTML]{206546}{-14.80}
& \underline{61.50}\tiny\textcolor{red}{+0.12} & \underline{45.01}\tiny\textcolor{red}{+4.24} & \underline{61.51}\tiny\textcolor{red}{+7.20} & 50.18\tiny\textcolor{red}{+13.56} & \underline{49.86}\tiny\textcolor{red}{+7.83} & \underline{17.53}\tiny\textcolor{red}{+6.01} & 59.85\tiny\textcolor[HTML]{206546}{-2.38} & \underline{55.39}\tiny\textcolor{red}{+6.09} \\

\arrayrulecolor{black!20}\midrule
\rowcolor{mycell}
Deepseek-R1 & \textbf{70.84} & \textbf{67.12} & \textbf{60.64} & \textbf{45.50} & \textbf{72.40}
& \textbf{76.63} & \textbf{58.17} & \textbf{67.30} & \textbf{59.69} & \textbf{61.16} & \textbf{24.37} & \textbf{67.70} & \textbf{64.02}\\

\arrayrulecolor{black}\bottomrule
\end{tabular}
}
\label{tab:qa-results}
\end{table*}

\paragraph{Question-Answering} 
We design the below eight challenging tasks using the Question-Answering format:
\textbf{\textit{(i)}} \textit{Entity-based Dynasty Determination} (\textbf{EDD}): infer the historical dynasty of a given entity based on contextual information.
\textbf{\textit{(ii)}} \textit{Plausibility Judgment} (\textbf{PJ}): assess whether a described historical scenario is plausible by reasoning about temporal and factual consistency.
\textbf{\textit{(iii)}} \textit{Temporal Order Understanding} (\textbf{TOU}): understand and compare the chronological order of historical events or figures.
\textbf{\textit{(iv)}} \textit{Relation Reasoning} (\textbf{RR}): reason about the historical relationships between entities, such as their spatial, temporal, or functional connections.
\textbf{\textit{(v)}} \textit{Script Error Correction} (\textbf{SEC}): identify and correct historical inaccuracies in visual or textual narratives.
\textbf{\textit{(v)}} \textit{Entity Evolution Understanding} (\textbf{EEU}): track and understand the evolution of entity names or attributes across different historical periods.
\textbf{\textit{(vi)}} \textit{Time Interval Calculation} (\textbf{TIC}): calculate the temporal gap between historical entities or events.
\textbf{\textit{(vii)}} \textit{Temporal Entity Selection} (\textbf{TES}): select the correct historical entity based on temporal and contextual constraints.
\textbf{\textit{viii}} \textit{Long Script Error Correction} (\textbf{LSEC}): identify and correct complex historical inaccuracies in long narratives by reasoning across extended contexts.
The key aspect of these task designs is to examine LLM’s ability to accurately \textbf{perceive and reason} about temporal relationships in a structured manner.\footnote{Each task's examples are presented in App.~\ref{app:qa-case}.}

\paragraph{Timeline Ito Game}
Our developed Timeline Ito Game is a collaborative reasoning game where agents infer the chronological order of historical entities within a dynasty timeline using thematic metaphors.
As shown in Figure~\ref{fig:intro}, the rules can be divided into the following steps:
\begin{itemize}
    \item \textbf{Step1: Describe Card}: Agents describe their assigned historical entity using a given theme without explicit temporal references.
    \item \textbf{Step2: Infer Rank}: Agents collaboratively deduce their relative positions in the timeline based on shared contexts.
    \item \textbf{Step3: Determine Order}: Each Agent sequentially predicts their position in the timeline relative to the others, and the team’s final order is based on these individual predictions.
\end{itemize}
The game ends when the team’s predicted order matches the true chronological sequence or when the maximum number of rounds, $K$, is reached.\footnote{We present a running case in App.~\ref{app:ito-case}.}
% If incorrect, a new theme is introduced, and the game continues to the next round.
% We present a running case in Appendix~\ref{app:ito-case} to enhance the understanding of the rules of this game.
% Through the developed rules, we can evaluate the temporal alignment capability of LLMs.

\subsection{Data Collection}
% To make the annotation process cost-efficient and
% accurate, we employ a two-stage funnel annotation
% strategy, combining LLM-based and human annotation. 
% In the first stage, GPT-4o~\citep{},
% performs initial annotations, followed by a second stage, where crowd-sourced human annotators conduct quality control. 

\begin{figure}[t]
    \centering
    \includegraphics[width=0.36\textwidth]{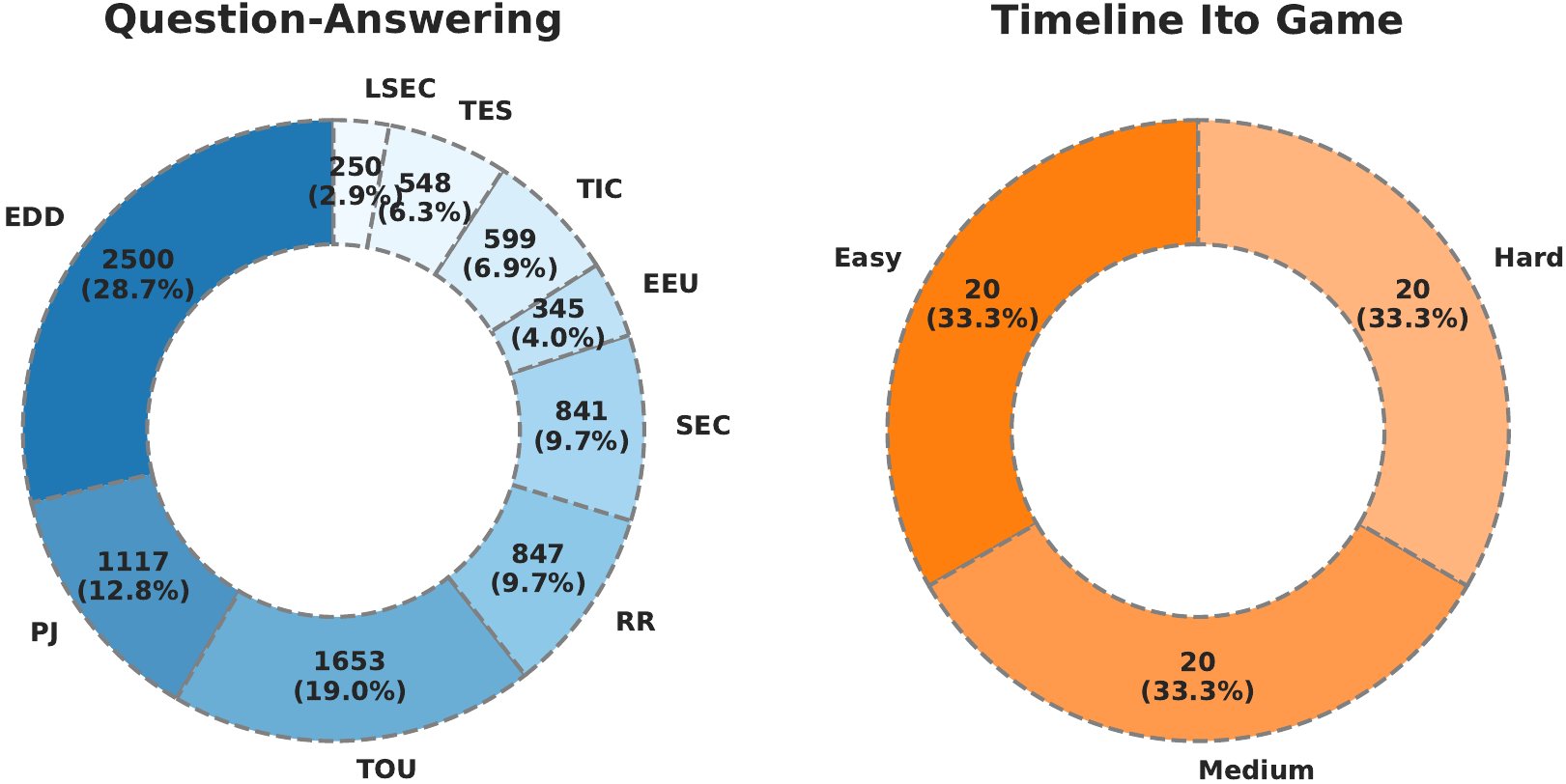}
    \caption{Statistic of CTM.
    }
    \label{fig:sta}
    \vspace{-5mm}
\end{figure}

\paragraph{Source}
We construct a comprehensive entity information repository by collecting diverse data from multiple authoritative sources, \textit{e.g.}, \texttt{Gushiwen}
% \footnote{\url{https://www.gushiwen.cn/}}
, \texttt{CBDB}
% \footnote{\url{https://projects.iq.harvard.edu/chinesecbdb}}
, \texttt{CHGIS}
% \footnote{\url{https://gis.harvard.edu/china-historical-gis}}
, \texttt{Wikipedia}
% \footnote{\url{https://zh.wikipedia.org/wiki/}}
, and 
\texttt{Ihchina}
% \footnote{\url{https://www.ihchina.cn/}}
.
The historical dynasties are simplified into ten major periods based on \texttt{Allhistory} and \texttt{CHINA—Timeline of Historical Periods}, 
specifically: ``先秦'', ``汉'', ``六朝'', ``隋'', ``唐'', ``五代'', ``宋'', ``元'',  ``明', ``清''.
The entity repository contains 1,652 figures (with attributes such as birth address, birth year, death year, and associated books or sentences), 2,907 places (including 990 primary administrative regions and 1,917 subordinate localities), 93 allusions, 49 ingredients, and 44 intangible cultural heritage items.

\begin{figure}[t]
    \centering
    \includegraphics[width=0.38\textwidth]{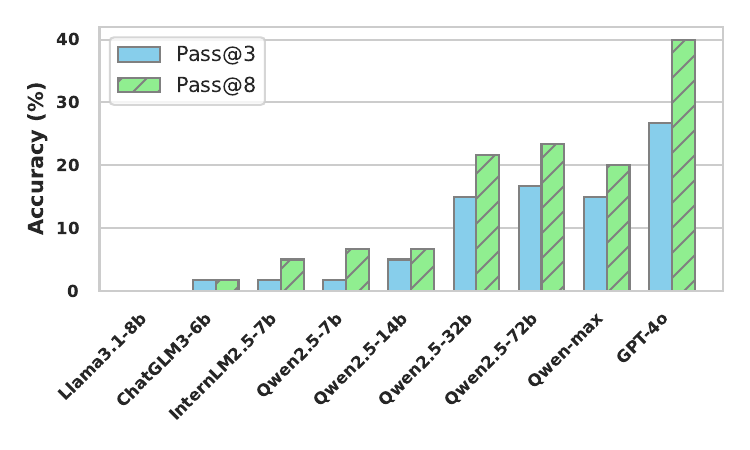}
    \caption{Average performance of Ito's Guessing Game. Detailed results can be found in Appendix~\ref{app:ito_acc}.
    }
    \label{fig:ito}
    \vspace{-5mm}
\end{figure}

\paragraph{Annotation Process}
The annotation process is structured into three key steps to ensure systematic and high-quality data generation:
\textbf{seed prompt creation}, \textbf{entity-aware data generation}, and \textbf{validation and quality control}.\footnote{The details of each step are provided in the App.~\ref{app:anno}.}
The process systematically generates annotated data while aligning with the repository's structured knowledge.
The statistics of CTM on the task are shown in Figure~\ref{fig:sta}.

\begin{figure*}[t]
    \centering
    \includegraphics[width=1\textwidth]{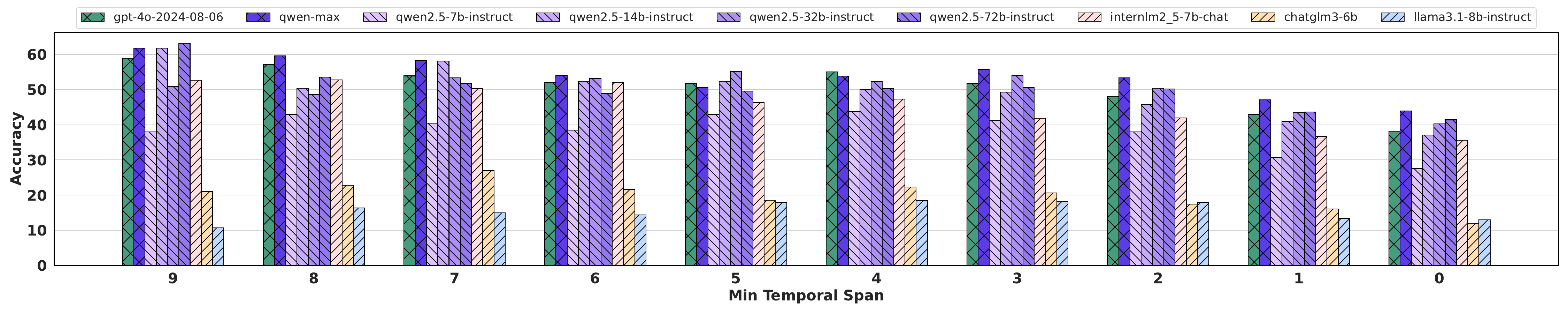}
    \caption{Accuracy across entity inter-dynastic intervals under direct prompting setting. 
    The detailed results are shown in Figure~\ref{fig:acc_span_cot}, Figure~\ref{fig:line_direct} and Figure~\ref{fig:line_cot}.
    }
    \label{fig:acc_span}
\end{figure*}

\subsection{Evaluation}
We use the \textbf{accuracy} metric to evaluate the QA tasks while \textbf{Pass@$K$} is used to evaluate Ito's Guessing Game.
Due to the varying lengths of LLM-generated
text, it is challenging to perform exact match evaluation.
We use GPT-4o as the evaluator\footnote{The prompt for the evaluator is provided in Appendix~\ref{app:prompt}.}, which determines the correctness of responses by comparing the prediction with the ground truth using the CoT~\cite{wei2022chain}.
Pass@$K$ measures whether the sequential alignment is achieved within $K$ attempts, we set $K$ to 3 and 8.

\section{Experiments}
\paragraph{Backbones}
We evaluate \textbf{twelve} mainstreaming LLMs,
the complete list of models is in App.~\ref{app:backbone_list}.
\subsection{Main Results}
Table~\ref{tab:qa-results} and Figure~\ref{tab:qa-results} present the experimental results of QA and Ito's Guessing Game, respectively.
We observe the following empirical findings:
\textbf{(I)} The more entities considered, the worse the performance, and Time Interval Calculation (TIC) is the most challenging task.
The former requires identifying the temporal information of multiple entities, while the latter demands a more precise assessment of specific timestamps.
\textbf{(II)} CoT can enhance performance, however, when the LLM is very small or the context is excessively long, it can even negatively impact temporal reasoning tasks.
This aligns with the conclusions of work~\cite{chu-etal-2024-timebench} and may be attributed to the knowledge sensitivity inherent in temporal reasoning.
\textbf{(III)} InternLM2.5 demonstrates strong performance among small open-source models, which may be attributed to the quality and composition of its training data.
\textbf{(IV)} The reasoning model demonstrates remarkably strong performance.
\textbf{(V)} Temporal alignment is highly challenging, and even powerful model GPT-4o fail to exceed 40 on the Pass@8 metric.
\textbf{(VI)} Small LLMs cannot align entities across different dimensions, and the Pass@$K$ performance for LLMs smaller than 32B does not exceed 10.

\subsection{Analysis}
\paragraph{The shorter the time interval between the entities, the greater the difficulty.}
As illustrated in Figure~\ref{fig:acc_span}, we evaluate performance across various models based on entity inter-dynastic intervals.
For example, an interval of 1 indicates adjacent dynasties, while an interval of 0 represents the same dynasty. 
As the interval decreases, performance declines.
This is because reasoning in QA tasks requires a clear understanding of the temporal relationships between entities, with closer intervals demanding more precise examination.

\begin{figure}[t]
    \centering
\includegraphics[width=0.40\textwidth]{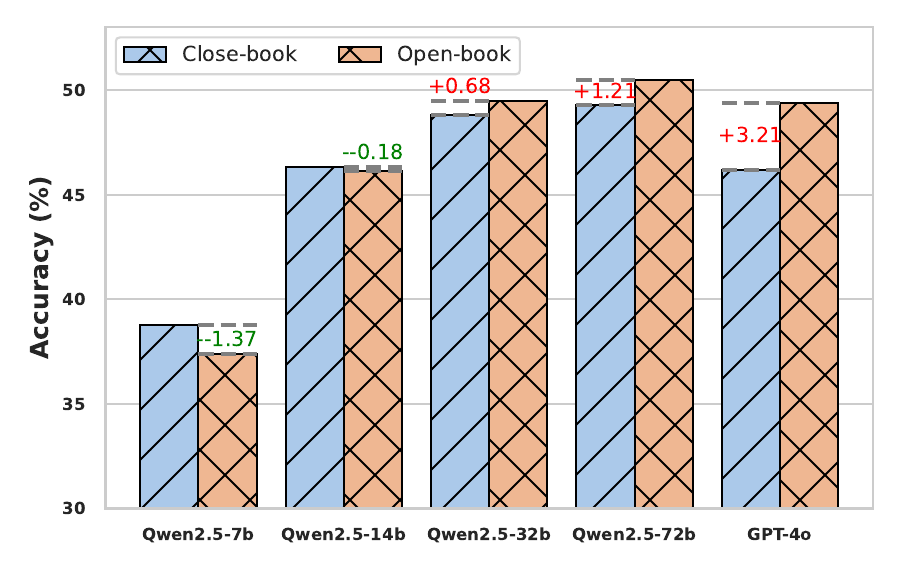}
    \caption{Performance in the close-book and open-book settings. Detailed results can be found in App.~\ref{app:openbook}.
    }
    \label{fig:openbook}
\end{figure}

\paragraph{In the open-book setting, temporal reasoning performance can be moderately improved.}
To obtain more precise temporal information about entities, we can leverage search engines to retrieve relevant information from the web, enhancing the specificity of entity details~\cite{wu2025webwalker}.
In the open-book setting, we use the titles and snippets of the Top-10 webpages retrieved via \texttt{Google} search as retrieval-augmented information.
As shown in Figure~\ref{fig:openbook}, it can be observed that performance improves after incorporating the retrieved content, except for Qwen2.5-7B, possibly due to its weaker longe contextual understanding.

\section{Conclusion}
We introduce CTM, a benchmark designed to
evaluate LLMs on temporal reasoning and alignment across Chinese dynasties.
CTM benchmark emphasizes contextualization, cross-entity relationships, and temporal alignment.
Empirical evaluations on various LLMs reveal the challenges posed by CTM, demonstrating that existing LLMs struggle with nuanced temporal understanding. 
These findings through analysis suggest the need for improved pretraining, structured knowledge integration, and refined reasoning mechanisms.
CTM provides a culturally rich resource for advancing temporal reasoning research. 

\section*{Limitations}

\paragraph{Prompt Design and Evaluation Settings}
This study evaluates the performance of LLMs on CTM using various prompts, including the most common settings of direct prompting and chain-of-thought (CoT).
However, it is acknowledged that the effectiveness of these prompts may vary across different tasks and models.
Future work could explore the possibility of dynamically adapting prompt designs to better suit specific temporal reasoning tasks, as well as expanding to more diverse few-shot and zero-shot settings.
As LLMs continue to evolve, it will be crucial to periodically update prompt strategies to ensure a robust and comprehensive evaluation.

\paragraph{Dataset Scale and Coverage}
While CTM currently includes a diverse range of Chinese temporal reasoning tasks, there is significant potential for expanding both its size and coverage.
With 8,750 examples already developed, the dataset can be further enriched with larger and more complex temporal scenarios, as well as longer historical events and a broader range of question types.
Additionally, the timeline Ito game data could be expanded to incorporate more intricate details and interesting themes, providing greater challenges for models and revealing their strengths and limitations.

% Bibliography entries for the entire Anthology, followed by custom entries
%\bibliography{anthology,custom}
% Custom bibliography entries only
\bibliography{custom}
\clearpage
\newpage
\appendix
\section{Related Works}
\label{app:rw}
\paragraph{Chinese Cultural Understanding in LLMs}
Recent advancements in LLMs have shown promise in cultural understanding tasks, with some studies specifically evaluating their performance in Chinese culture, including assessments of commonsense knowledge~\cite{shi2024corecode,sun-etal-2024-benchmarking-chinese,li-etal-2024-cmmlu}, foodie culture~\cite{li-etal-2024-foodieqa}, and historical knowledge~\cite{bai2024baijia}.
As one of the world's longest-standing cultures, Chinese culture spans a vast historical timeline, with each dynasty rich in historical figures, anecdotes, and cultural narratives.
Its strong cultural attributes also allow for effective contextualization. 
This makes dynastic timelines particularly well-suited for temporal reasoning and alignment in our work.

\paragraph{Temporal Reasoning in LLMs}
Temporal reasoning is a critical capability for LLMs, with existing benchmarks focusing on factual temporal grounding~\citep{chen2021a, dhingra-etal-2022-time}, complex temporal logic~\citep{tan-etal-2023-towards,su-etal-2024-living}, and multi-granular temporal awareness \citep{chu-etal-2024-timebench,islakoglu2025chronosense}. 
As shown in Table~\ref{table:comparsion}, these benchmarks are primarily English-based and rely on rule-based dataset construction, which limits contextualization and diversity. 
In contrast, \textbf{CTM} is grounded in Chinese culture and leverages LLM-based question generation, resulting in more flexible and contextually relevant questions. 
Our benchmark also features a broader range of tasks to access the reasoning and alignment of LLMs and ensures more accurate entity assessments that require nuanced temporal understanding.

\section{English Translations}
\subsection{Ten Major Dynasties and Corresponding Period}
``先秦 (Pre-Qin)'' (-2100 to -206), ``汉 (Han)'' (-206 to 220), ``六朝'' (Six Dynasties) (220 to 589), ``隋（Sui）'' (581 to 618), ``唐 (Tang)'' (618 to 906), ``五代 (Five Dynasties)'' (907 to 960), ``宋 (Song)'' (960 to 1279), ``元 (Yuan)'' (1279 to 1368),  ``明 (Ming)'' (1368 to 1644), ``清 (Qing)'' (1644 to 1912).

\subsection{Translated QA Pair}
\label{app:tran_qa}
\begin{itemize}
    \item[\textbf{Q.}] 以下是一段镜头描述，其中有一处或多处不遵循真实历史背景的穿帮内容，请指出:  
    
    The following is a scene description containing one or more anachronisms that do not align with historical accuracy. Please identify them:  
    
    李白在创作《将进酒》时，白居易在一旁吟诗，同时桌上摆着一盘辣椒，旁边还有一位乐师在演奏古琴。  
    
    While \textbf{Li Bai} is composing \textit{``Bring in the Wine''}, \textbf{Bai Juyi} is reciting poetry beside him. On the table, there is a plate of \textbf{chili peppers}, and a musician is playing the \textbf{guqin art} nearby.

    \item[\textbf{A.}] 穿帮内容:
    
    Anachronisms:  
    
    1. 李白去世时（762 CE），白居易还未出生（772 CE），两人不可能同时在场。
    
    When \textbf{Li Bai} passed away (762 CE), \textbf{Bai Juyi} had not yet been born (772 CE), making it impossible for them to be present together.
        
    2. 古琴艺术在唐朝时已非常成熟，符合历史背景。  
    
    \textbf{Guqin Art} was already well-developed during the Tang Dynasty, which aligns with the historical context.
        
    3. 辣椒在明朝才传入中国，不可能出现在唐朝。 
    
    \textbf{Chili peppers} were not introduced to China until the Ming Dynasty, so they could not have appeared during the Tang Dynasty.
\end{itemize}

\section{Statistics of CTM}
The statistics of CTM on tasks are shown in Table~\ref{tab:sta}.
% \clearpage
% \newpage
\begin{table}[ht]
\centering
\caption{ 
The statistics of CTM.}
\tiny
\setlength\tabcolsep{6pt}
\resizebox{0.5\textwidth}{!}{%
\begin{tabular}{c|c|c|c|c|c|c|c|c|c}
\toprule
\multicolumn{1}{c|}{\multirow{3}{*}{\textbf{Statistic}}} & \multicolumn{9}{c}{\textbf{Question-Answering}}\\
\cmidrule{2-10}
& \textbf{EDD} & \textbf{PJ} & \textbf{TOU} & \textbf{RR} & \textbf{SEC} & \textbf{EEU} & \textbf{TIC} & \textbf{TES} & \textbf{LSEC}\\
\midrule
\# Sample & 2500 & 1117 & 1653 & 847 & 841 & 345 & 599 & 548 & 250\\
\midrule
Cross Temp Count & 1 & \multicolumn{7}{c|}{2, 3, 4..10} & 4..15\\
\midrule
\midrule
\multicolumn{1}{c|}{\multirow{3}{*}{\textbf{Statistic}}} & \multicolumn{9}{c}{\textbf{Timeline Ito Game}}\\
\cmidrule{2-10}
& \multicolumn{3}{c|}{\textbf{Easy}} & \multicolumn{3}{c|}{\textbf{Medium}} & \multicolumn{3}{c}{\textbf{Hard}}\\
\midrule
\# Sample & \multicolumn{3}{c|}{20} & \multicolumn{3}{c|}{20} & \multicolumn{3}{c}{20}\\
\midrule
Cross Temp Count & \multicolumn{3}{c|}{3} & \multicolumn{3}{c|}{4} & \multicolumn{3}{c}{5} \\
\midrule
Agent Num & \multicolumn{3}{c|}{3} & \multicolumn{3}{c|}{4} & \multicolumn{3}{c}{5} \\
\bottomrule
\end{tabular}
}
\label{tab:sta}
\end{table}

\section{Entity Repository}
Figure~\ref{fig:figure_case}, Figure~\ref{fig:place_case}, Figure~\ref{fig:event_case}, Figure~\ref{fig:ingredient_case} and Figure~\ref{fig:ih_case} show the case of historical figure, place, event, ingredient and intangible cultural heritage, respectively.

\begin{figure}[ht]
\centering
\begin{adjustbox}{width=0.33\textwidth}
\begin{tcolorbox}[title={\textbf{\small Historical Figure}}, colback=whitesmoke, colframe=gray, boxrule=2pt, arc=0mm, width=0.5\textwidth]
{\small
\begin{verbatim}
## JSON Format
{
    "屈原": {
        "dynasty": "先秦",
        "address": "楚国丹阳秭归（今湖北宜昌）",
        "year_birth": "-340",
        "year_death": "-278",
        "book_and_sentences": [
            {
                "sentence": "身既死兮神以灵，魂魄毅兮为鬼雄。",
                "book": "《国殇》"
            },
            {
                "sentence": "路曼曼其修远兮，吾将上下而求索。",
                "book": "《离骚》"
            },
            # ...
        ]
    },
    # ...
}
\end{verbatim}
}
\end{tcolorbox}
\end{adjustbox}
\caption{A JSON-format case for historical figure entity.}
\label{fig:figure_case}
\end{figure}

\vspace{2cm}

\begin{figure}[H]
\centering
\begin{adjustbox}{width=0.33\textwidth}
\centering
\begin{tcolorbox}[title={\textbf{\small Place}}, colback=whitesmoke, colframe=gray, boxrule=2pt, arc=0mm]
{\small
\begin{verbatim}
## JSON Format
{
  "巴州": {
    "dynasty": "唐",
    "id": "hvd_111423",
    "begin": "758",
    "end": "762",
    "pre_address": "四川省巴中市",
    "subordinate_units": [
      {
        "begin_year": "758",
        "end_year": "762",
        "child_id": "hvd_44640",
        "name": "七盘县",
        "pre_address": "今四川巴中县西北一百二十里（今旺苍县东南"
      },
      # ...
    ]
  }
  #...
}
\end{verbatim}
}
\end{tcolorbox}
\end{adjustbox}
\caption{A JSON-format case for place entity.}
\label{fig:place_case}
\end{figure}

\begin{figure}[H]
\centering
\begin{adjustbox}{width=0.33\textwidth}
\begin{tcolorbox}[title={\textbf{\small Event}}, colback=whitesmoke, colframe=gray, boxrule=2pt, arc=0mm]
{\small
\begin{verbatim}
## JSON Format
{
  "司马迁写《史记》": {
    "id": "070",
    "dynasty": "汉",
    "main_figures": "司马迁"
  },
  # ...
}
\end{verbatim}
}
\end{tcolorbox}
\end{adjustbox}
\caption{A JSON-format case for event entity.}
\label{fig:event_case}
\end{figure}

\begin{figure}[H]
\centering
\begin{adjustbox}{width=0.33\textwidth}
\begin{tcolorbox}[title={\textbf{\small Ingredient}}, colback=whitesmoke, colframe=gray, boxrule=2pt, arc=0mm]
{\small
\begin{verbatim}
## JSON Format
{
  "水稻": {
    "dynasty": "先秦",
    "origin": "中国"
  },
  # ...
}
\end{verbatim}
}
\end{tcolorbox}
\end{adjustbox}
\caption{A JSON-format case for ingredient entity.}
\label{fig:ingredient_case}
\end{figure}

\begin{figure}[H]
\centering
\begin{adjustbox}{width=0.33\textwidth}
\begin{tcolorbox}[title={\textbf{\small Intangible Cultural Heritage}}, colback=whitesmoke, colframe=gray, boxrule=2pt, arc=0mm]
{\small
\begin{verbatim}
## JSON Format
{
  "昆曲": {
    "dynasty": "明",
    "place": "北京, 上海, 江苏省, 浙江省, 湖南省",
    "type": "表演艺术"
  },
  # ...
}
\end{verbatim}
}
\end{tcolorbox}
\end{adjustbox}
\caption{A JSON-format case for intangible cultural heritage entity.}
\label{fig:ih_case}
\end{figure}

\section{Annotation}
\label{app:anno}
\begin{itemize}
    \item \textbf{Step1: Seed Prompt Creation}: For each entity type, we manually design seed prompts~\citep{alpaca} to guide the self-instruct-based data generation process. 
    These prompts serve as templates to ensure diversity and relevance in the generated data.
    \item \textbf{Step2: Entity-Aware Data Generation}:  During LLM-based generation, the LLMs dynamically incorporate entity descriptions sampled from the pre-constructed entity information repository. This ensures that the generated content is contextually grounded in the repository's structured knowledge, enhancing control over entity-related information.
    \item \textbf{Step3: Validation and Quality Control}: After generation, each data point undergoes a validation step, where the temporal entities \textbf{mentioned} in the output are cross-referenced with the repository. 
    This ensures the accuracy and consistency of the entities, aligning the generated data with the repository's constraints.
\end{itemize}

\section{Cases}\label{app:cases}
\subsection{Cases in Question-Answering}\label{app:qa-case}
Figure~\ref{fig:EDD_case}, Figure~\ref{fig:PJ_case}, Figure~\ref{fig:TOU_case}, Figure~\ref{fig:RR_case}, Figure~\ref{fig:EEU_case}, Figure~\ref{fig:SEC_case}, Figure~\ref{fig:TIC_case}, Figure~\ref{fig:TES_case}, and Figure~\ref{fig:LSEC_case} show the Entity-based Dynasty Determination, Plausibility Judgment, Temporal Order Understanding, Relation Reasoning, Script Error Correction, Entity Evolution Understanding, Time Interval Calculation, Temporal Entity Selection and Long Script Error Correction tasks in JSON-format, respectively.

\begin{figure}[ht]
\begin{tcolorbox}[title={\textbf{\small Entity-based Dynasty Determination (EDD)}}, colback=whitesmoke, colframe=royalblue(web), boxrule=2pt, arc=0mm]
{\small
\begin{verbatim}
## JSON Format
{
  "type": "根据食材确定传入朝代",
  "question": "假设你穿越到某个朝代，听商贩说辣椒是最近几年才有的食材，你能推测你穿越到了哪个朝代吗？",
  "temporal_entities": ["辣椒"],
  "construct_explanation": "这个问题通过穿越情境让提问者推测辣椒的引入时间。辣椒最早于明朝从美洲传入中国，因此问题涉及1个时空实体-朝代关系，辣椒传入中国的朝代是‘明’。",
  "answer": "明"
}
\end{verbatim}
}
\end{tcolorbox}
\caption{A JSON-format case in EDD type of QA.}
\label{fig:EDD_case}
\end{figure}

\begin{figure}[ht]
\begin{tcolorbox}[title={\textbf{\small Plausibility Judgment (PJ)}}, colback=whitesmoke, colframe=royalblue(web), boxrule=2pt, arc=0mm]
{\small
\begin{verbatim}
## JSON Format
{
  "type": "合理性判断", 
  "question": "苏轼是眉州人，眉州是今四川眉山市，而四川以麻辣闻名。那么，苏轼是否吃辣？",
  "temporal_entities": ["苏轼", "眉州", "辣椒"],
  "construct_explanation": "苏轼生活在宋代（约公元1037年-1101年），辣椒最早于明朝传入中国。因此问题涉及3个时空实体-朝代关系，‘苏轼-宋’、‘眉州-宋’和‘辣椒-明’，通过时间线推理，苏轼生活的时代不可能接触到辣椒。",
  "answer": "否"
}
\end{verbatim}
}
\end{tcolorbox}
\caption{A JSON-format case in PJ type of QA.}
\label{fig:PJ_case}
\end{figure}

\begin{figure}[H]
\begin{tcolorbox}[title={\textbf{\small Temporal Order Understanding (TOU)}}, colback=whitesmoke, colframe=royalblue(web), boxrule=2pt, arc=0mm]
{\small
\begin{verbatim}
## JSON Format
{
  "type": "时间顺序理解",
  "question": "请将以下实体按时间顺序排列：屈原，李白，白居易，辣椒，古琴艺术，石榴？",
  "temporal_entities": ["屈原", "李白", "白居易", "辣椒", "古琴艺术", "石榴"],
  "construct_explanation": "屈原生活在先秦（约公元前340年-公元前278年），李白生活在唐代（约公元701年-762年），白居易生活在唐代（约公元772年-846年），辣椒在明朝传入中国（约公元16世纪），古琴艺术起源于先秦（约公元前11世纪），石榴在汉代传入中国（约公元前2世纪）。因此问题涉及6个时空实体-朝代关系，‘屈原-先秦’、‘李白-唐’、‘白居易-唐’、‘辣椒-明’、‘古琴艺术-先秦’和‘石榴-汉’，按时间顺序排列为古琴艺术、屈原、石榴、李白、白居易、辣椒。",
  "answer": "古琴艺术、屈原、石榴、李白、白居易、辣椒"
}
\end{verbatim}
}
\end{tcolorbox}
\caption{A JSON-format case in TOU type of QA.}
\label{fig:TOU_case}
\end{figure}

\begin{figure}[H]
\begin{tcolorbox}[title={\textbf{\small Relation Reasoning (RR)}}, colback=whitesmoke, colframe=royalblue(web), boxrule=2pt, arc=0mm]
{\small
\begin{verbatim}
## JSON Format
{
  "type": "关系判断",
  "question": "虞州和河东郡在历史上有什么关系？",
  "temporal_entities": ["虞州", "河东郡"],
  "construct_explanation": "虞州和河东郡在历史上的行政区划大致对应现在的山西运城。河东郡是古代的一个重要行政区划，虞州则是后来的行政区划。因此问题涉及2个时空实体-朝代关系，‘虞州-隋’和‘河东郡-先秦’，两者在地理位置上与现代的山西运城有较高的重合度。",
  "answer": "都为现山西运城"
}
\end{verbatim}
}
\end{tcolorbox}
\caption{A JSON-format case in RR type of QA.}
\label{fig:RR_case}
\end{figure}

\begin{figure}[t]
\begin{tcolorbox}[title={\textbf{\small Entity Evolution Understanding (EEU)}}, colback=whitesmoke, colframe=royalblue(web), boxrule=2pt, arc=0mm]
{\small
\begin{verbatim}
## JSON Format
{
  "type": "实体名称变迁",
  "question": "在唐朝的浈州内，今日的四会县在当时名为何？",
  "temporal_entities": ["浈州", "四会县"],
  "construct_explanation": "唐朝时期存在的浈州（公元634年至638年）包含今日的四会县。根据历史资料，此时四会县作为浈州的下属地方，在唐朝已经是现名四会县。因此问题涉及2个时空实体-地方名称关系，‘浈州-唐’ 及‘四会县-唐’，从历史上来看，当时的四会县没有变更过名称。",
  "answer": "四会县"
}
\end{verbatim}
}
\end{tcolorbox}
\caption{A JSON-format case in EEU type of QA.}
\label{fig:EEU_case}
\end{figure}

\begin{figure}[t]
\begin{tcolorbox}[title={\textbf{\small Script Error Correction (SEC)}}, colback=whitesmoke, colframe=royalblue(web), boxrule=2pt, arc=0mm]
{\small
\begin{verbatim}
## JSON Format
{
  "type": "穿帮镜头指正",
  "question": "以下是一段镜头描述，其中有一处不遵循真实历史背景的穿帮内容，请指出: 在唐代的朗州，一位表演者正在用昆曲演绎盛唐的繁华，而高粱酒则是当场招待贵宾的饮品。",
  "temporal_entities": ["昆曲", "朗州", "高粱"],
  "construct_explanation": "1. 朗州存在于唐代从621年到741年之间。2. 昆曲起源于明代，与唐代不重叠。3. 高粱传入中国的时间更为久远，早在先秦时期便已存在，因此可用于唐代饮品制作。因此，使用昆曲演绎显然是历史穿帮。",
  "answer": "昆曲在唐代的朗州出现是穿帮内容，因昆曲起源于明代。"
}
\end{verbatim}
}
\end{tcolorbox}
\caption{A JSON-format case in SEC type of QA.}
\label{fig:SEC_case}
\end{figure}

\begin{figure}[H]
\begin{tcolorbox}[title={\textbf{\small Time Interval Calculation (TIC)}}, colback=whitesmoke, colframe=royalblue(web), boxrule=2pt, arc=0mm]
{\small
\begin{verbatim}
## JSON Format
{
  "type": "时间差计算",
  "question": "从屈原投江到李白出生，再到苏轼出生，中间经历了多少年？",
  "temporal_entities": ["屈原", "李白", "苏轼"],
  "construct_explanation": "屈原投江发生在约公元前278年，李白出生于公元701年，苏轼出生于公元1037年。因此问题涉及3个时空实体-朝代关系，‘屈原-先秦’、‘李白-唐’和‘苏轼-宋’，通过计算时间差，屈原到李白相差约979年，李白到苏轼相差约336年，总共相差约1315年。",
  "answer": "约1315年"
}
\end{verbatim}
}
\end{tcolorbox}
\caption{A JSON-format case in TIC type of QA.}
\label{fig:TIC_case}
\end{figure}

\begin{figure}[H]
\begin{tcolorbox}[title={\textbf{\small Temporal Entity Selection (TES)}}, colback=whitesmoke, colframe=royalblue(web), boxrule=2pt, arc=0mm]
{\small
\begin{verbatim}
## JSON Format
{
  "type": "选出对应时空实体",
  "question": "以下四个实体中，哪个属于唐朝？\n(A) 李白 \n(B) 苏轼 \n(C) 屈原 \n(D) 曹操",
  "temporal_entities": ["李白", "苏轼", "屈原", "曹操"],
  "construct_explanation": "李白生活在唐代（约公元701年-762年），苏轼生活在宋代（约公元1037年-1101年），屈原生活在先秦（约公元前340年-公元前278年），曹操生活在汉代（约公元155年-220年）。因此问题涉及4个时空实体-朝代关系，‘李白-唐’、‘苏轼-宋’、‘屈原-先秦’和‘曹操-汉’，通过朝代背景推理，李白属于唐朝。",
  "answer": "李白"
}
\end{verbatim}
}
\end{tcolorbox}
\caption{A JSON-format case in TES type of QA.}
\label{fig:TES_case}
\end{figure}

\begin{figure*}[t]
\begin{tcolorbox}[title={\textbf{\small Long Script Error Correction (LSEC)}}, colback=whitesmoke, colframe=royalblue(web), boxrule=2pt, arc=0mm]
{\small
\begin{verbatim}
## JSON Format
{  "type": "剧本穿帮问题指正_长上下文", 
  "question": "以下是一段剧本描述，请指出其中不符合历史背景的穿帮内容：\n\n背景设定在东汉末年，名将曹操正在洛阳的一座文人聚会酒楼中与手下谋士们讨论政治大计。酒楼内弥漫着美酒的香气，桌上的菜肴色香俱全，包括蜜饯杏仁、红烧鹿肉以及一盘山药炖鸡。曹操端起酒杯，起身与众人敬酒，话题转向了当前国家的局势。突然，酒楼的门外传来一阵欢声笑语，几位文人相伴走入酒楼，他们是当时知名的诗人王之涣和杜牧。王之涣手捧一卷《登鹳雀楼》的诗稿，热衷地与杜牧讨论诗句的对仗工整。两位诗人落座后，曹操也邀请他们一起品酒作诗。\n\n就在此时，一位穿着朴素的商人走了进来，自称是商贸使者张衡。他向酒楼的主人要求提供一些来自西南的特产，并推荐了其中的一种热带水果——草莓。酒楼老板随即呈上了这一新鲜水果，几人尝过后纷纷表示口感非常独特，直言这是一种他们从未品尝过的美味。与此同时，另一位年轻的学者走进酒楼，他正是历史学者班固，他从洛阳的学府赶来，听到酒楼内的喧闹声，不禁走入与众人交谈。\n\n随着酒楼里的谈话逐渐深入，众人讨论的议题转向了国家的未来。王之涣对曹操表示，若有朝一日可以恢复汉朝的光辉，他愿意为之创作更多诗篇。而杜牧则提到，诗词创作对于平定民心具有重要作用。他们的对话引发了在座每个人的深思。\n\n此时，酒楼外突然传来一阵马蹄声，一位身穿铠甲的将军匆匆走进酒楼，向曹操报告前线战况。这位将军正是夏侯惇，他刚从许都赶来，带来了最新的军情。曹操听后眉头紧锁，立即召集众人商议对策。夏侯惇提议联合孙权共同对抗袁绍，但曹操却认为应当先稳固内部，再图外敌。\n\n就在众人争论不休时，酒楼的门再次被推开，一位身穿道袍的老者走了进来。他自称是华佗，手中捧着一瓶刚刚炼制的‘麻沸散’。华佗向曹操献上药物，声称服用后可缓解头痛。曹操接过药物，若有所思地看了看众人，随后将药物放入怀中。\n\n酒楼的氛围逐渐热烈起来，众人一边饮酒一边讨论着国家的未来。突然，一位年轻的乐师走到酒楼中央，开始演奏一曲《短歌行》。琴声悠扬，众人纷纷停下手中的酒杯，静静地聆听。乐师演奏完毕后，曹操起身鼓掌，称赞其技艺高超，并邀请他加入自己的幕府。\n\n随着夜幕降临，酒楼内的灯火逐渐点亮，众人继续畅谈。曹操举起酒杯，高声说道：‘今日与诸位相聚，实乃幸事！愿我们共同努力，恢复汉室荣光！’众人纷纷举杯响应，酒楼内充满了欢声笑语。",
  "temporal_entities": ["曹操", "王之涣", "杜牧", "张衡", "草莓", "班固", "夏侯惇", "华佗"],
  "construct_explanation": "剧本中出现的多个历史人物和食材存在不符合历史背景的穿帮内容：\n\n1. 王之涣和杜牧不可能同时出现在东汉末年。王之涣生活在唐朝，杜牧则生活在唐朝晚期，二人不可能与曹操同时存在。\n2. 张衡生活在东汉时期，他是著名的天文学家和文学家，但剧本中将他错误地设定为商贸使者，这一身份与历史背景不符。\n3. 草莓直到清朝才传入中国，剧本中的草莓出现在东汉末年不符合历史事实。\n4. 班固作为东汉时期的历史学者，他比曹操年长不少，因此不可能与曹操同时出现在剧本中。班固应当是年老之际，无法与年轻的曹操同时活跃。",
  "answer": "1. 王之涣和杜牧不可能同时出现在东汉末年。\n2. 张衡不可能作为商贸使者与曹操同时出现。\n3. 草莓在东汉末年未传入中国，属于不合时宜的食材。\n4. 班固的年纪和身份不应与曹操的同时期相关。"
}
\end{verbatim}
}
\end{tcolorbox}
\caption{A JSON-format case in LSEC type of QA.}
\label{fig:LSEC_case}
\end{figure*}

\subsection{Running Example of Timeline Ito Game}\label{app:ito-case}
% A running example is shown in Figure~\ref{fig:ito_public_memory_case}.
A Timeline Ito Game running example given the ``\textit{fruit size}'' theme is below.
\begin{figure}[H]
\begin{tcolorbox}[title={\textbf{\small A Running Example of Timeline Ito Game}}, colback=whitesmoke, colframe=royalblue(web), boxrule=2pt, arc=0mm]
{\small
\begin{verbatim}
-----------------------------------------
真实顺序：{"屈原": 1, "李白": 2, "苏轼": 3}
初始化：
Agent P1: "李白"
Agent P2: "屈原"
Agent P3: "苏轼"
-----------------------------------------
Agent P3 prediction in Round 1:
{
  "分析": "根据对xxx的了解，xxx是宋朝时期的一位著名诗人。因此，时间实体对应的朝代是宋。根据历史朝代时间线，宋朝在十个朝代中居中偏后的位置。结合水果大小的主题，应该选择稍大一些但不是最大的水果。在已提供的选择中，西瓜是最大的，蓝莓是最小的，对应最前的朝代。选择朝代在居中偏后的主题实体是桃子。",
  "理由": "我的时间实体对应的朝代居中偏后",
  "主题实体": "桃子"
}
-----------------------------------------
###=== Round 1 ===
当前主题: 水果大小
Agent P1: 我的时间实体对应的朝代居中偏前. 因此我选择橙子
Agent P2: 我的时间实体对应的朝代最前. 因此我选择蓝莓
Agent P3: 我的时间实体对应的朝代居中偏后. 因此我选择桃子
本轮排序结果: {'P1': 2, 'P2': 1, 'P3': 2}

本轮排序错误，错误的玩家：P3

###=== Round 2 ===
当前主题: 书本厚度
Agent P1: 我的时间实体对应的朝代居中. 因此我选择字典
Agent P2: 我的时间实体对应的朝代最前. 因此我选择书签
Agent P3: 我的时间实体对应的朝代居中偏后. 因此我选择百科全书

本轮排序结果: {'P1': 2, 'P2': 1, 'P3': 3}
游戏结束！所有玩家的排序正确。
\end{verbatim}
}
\end{tcolorbox}
% \caption{A Timeline Ito Game running example given the ``\textit{fruit size}'' theme.}
\label{fig:ito_public_memory_case}
\end{figure}

\section{LLM Backbone List}\label{app:backbone_list}
We validate the total number of 
twelve models, including both closed-sourced and open-sourced ones~\cite{achiam2023gpt,dubey2024llama,yang2024qwen2,cai2024internlm2,glm2024chatglm}.
The complete list of evaluated LLMs is shown in Table~\ref{tab:language_models}.
\begin{table}
    \centering
    \tiny
    \begin{tabular*}{0.45\textwidth}{@{\extracolsep{\fill}}lccc}
    \toprule
    \textbf{Models} & \textbf{Full Name} &\textbf{Open Source?} & \textbf{Model Size}\\
    \midrule
    GPT-4o &  gpt-4o-2024-08-06 & \no & -\\
    Qwen-max &  qwen-max & \no & -\\
    o1-preview & o1-preview & \no & -\\
    \midrule
    LLaMA3.1$_{\text{8b}}$ & Meta-Llama-3.1-8B-Instruct & \yes & 8B\\
    ChatGLM3$_{\text{6b}}$ & chatglm3-6b & \yes & 6B\\
    InternLM2.5$_{\text{7b}}$ & internlm2\_5-7b-chat & \yes & 7B\\
    Qwen2.5$_{\text{7b}}$ & qwen2.5-7b-instruct & \yes & 7B\\
    Qwen2.5$_{\text{14b}}$ & qwen2.5-14b-instruct & \yes & 14B\\
    Qwen2.5$_{\text{32b}}$ & qwen2.5-32b-instruct & \yes & 32B\\
    Qwen2.5$_{\text{72b}}$ & qwen2.5-14b-instruct & \yes & 72B\\
    DeepSeek-R1 & deepseek-r1 & \yes & 671B\\
    \bottomrule
    \end{tabular*}
    \normalsize
    \caption{LLMs evaluated in our experiments}
    \label{tab:language_models}
\end{table}

\section{Prompt}\label{app:prompt}

\begin{figure}[H]
\begin{tcolorbox}[title={\textbf{\small Prompt for Direct Prediction}}, colback=whitesmoke, colframe=lightred, boxrule=2pt, arc=0mm]
{\small
\begin{verbatim}
请回答以下问题：
{question}
\end{verbatim}
}
\end{tcolorbox}
\caption{Prompt for Direct Prediction}
\label{fig:direct_prompt}
\end{figure}

\begin{figure}[H]
\begin{tcolorbox}[title={\textbf{\small Prompt for CoT Prediction}}, colback=whitesmoke, colframe=lightred, boxrule=2pt, arc=0mm]
{\small
\begin{verbatim}
请按照以下步骤一步一步思考并回答问题：
1. {question}
2. 思考并详细分析问题，然后得到结论及简要理由，如果理由比较复杂请分条简要列出。
3. 请以 JSON 格式返回结果，格式如下：
{
  "思考": "请在这里填写详细的思考过程",
  "回答": "请在这里填写结论及理由"
}
\end{verbatim}
}
\end{tcolorbox}
\caption{Prompt for CoT Prediction.}
\label{fig:cot_prompt}
\end{figure}

\begin{figure*}[t]
\begin{tcolorbox}[title={\textbf{\small Prompt of Evaluator}}, colback=whitesmoke, colframe=lightred, boxrule=2pt, arc=0mm]
{\small
\begin{verbatim}
你是一个专业的问答系统评估员。请根据以下信息评估答案的质量，并输出详细的思考过程：

**问题类型**：{question_type}
**问题**：{question}
**参考答案**：{reference} {answer}
**待评估答案**：{prediction}

**评估步骤**：
1. 判断问题类型：
- 如果问题是“是否问题”，进入步骤 2。
- 如果问题不是“是否问题”，进入步骤 3。
2. 对于“是否问题”：
- 检查待评估答案是否正确回答了“是/否”。
- 检查待评估答案的原因是否与参考答案一致。
- 如果两者都正确，评估结果为1；否则评估结果为0。
3. 对于非“是否问题”：
- 检查待评估答案是否**完全覆盖**参考答案中的所有关键点。
- 检查待评估答案中的每个点是否与参考答案**完全一致**，包括事实、逻辑和时间线等。
- 检查朝代是否宽松匹配：
  - 比较待评估答案中的朝代范围与参考答案的朝代范围，允许评估答案使用细分的朝代划分。
  - 参考答案中的朝代范围：先秦（-2100～-206），汉（-206～220），六朝（220～589），隋（581～618），唐（618～906），五代（907～960），宋（960～1279），元（1279～1368），明（1368～1644），清（1644～1912）。
  - 秦朝以前的朝代范围夏、商、周、春秋、战国等都属于先秦。六朝和五代的朝代范围分别为东晋、宋、齐、梁、陈、后周和后梁、后唐、后晋、后汉、后周。
  - 允许待评估答案中的细分朝代在参考答案的朝代范围内进行宽松匹配。
- 如果待评估答案**完全覆盖且完全一致**，评估结果为1；否则返回评估结果为0。

**思考过程**：
- 详细分析待评估答案与参考答案的异同。
- 重点关注跨时空推理的历史时间线和逻辑一致性。
- 判断多个时空实体之间及与朝代之间的关系。

**请以 JSON 格式返回结果**：
{
  "思考": "请在这里填写详细的思考过程", 
  "评估结果": "请在这里填写0或1"
}
\end{verbatim}
}
\end{tcolorbox}
\caption{A JSON-format case in intangible cultural heritage entity.}
\label{fig:eval_prompt}
\end{figure*}

\begin{figure*}[t]
\begin{tcolorbox}[title={\textbf{\small Prompt For Step1 of Timeline Ito Game}}, colback=whitesmoke, colframe=lightred, boxrule=2pt, arc=0mm]
{\small
\begin{verbatim}
你是一个玩家，参与了一个叫做“命悬一线（ito）”的游戏，现在你收到一个时间实体：'{self.entity}'。
1. 首先需要推理得出该时间实体所属的朝代，朝代时间线为：
先秦（-2100～-206），汉（-206～220），六朝（220～589），隋（581～618），唐（618～906），五代（907～960），宋（960～1279），元（1279～1368），明（1368～1644），清（1644～1912）

2. 主题与朝代的对应规则：如果主题是”水果大小“，则水果越小，对应朝代越早；如果是“书本厚度”，则书本越薄，对应朝代越早；如果是“船只大小”，则船只越小，对应朝代越早；如果是“电子产品”大小，则电子产品越小，对应朝代越早；如果是“行星大小”，则行星越小，对应朝代越早；如果是动物大小，则动物越小，对应朝代越早；如果是“建筑高度”，则建筑越矮，对应朝代越早；如果是“水的温度”，则水越冷，对应朝代越早。
本轮的主题是{theme}，主题实体的个数与朝代个数对应。你需要在{theme_entities}中选择1个对应“主题实体在主题的顺序”，来代表你的“时间实体的朝代在朝代时间线中的位置”，返回你认为最能代表你的“时间实体的朝代”的1个“主题实体”

3. 以下是之前的交流记录，排序相关判断可作为参考，其余的不作参考。(如果这里是空的，就不参考)
{public_memory_str}

4. 分析你的时间实体在朝代时间线中的位置，输出分析过程，分析内容中你的“时间实体”和你推理出的“朝代”用“xxx”代替，只能出现“主题实体”的名称，可以用“朝代偏前或偏后”等表达。请注意，你的分析过程需要符合主题与朝代的对应规则。
结合分析给出一句话表达你的推理理由，“我的时间实体对应的朝代最前/偏前/居中/偏后/最后等”（理由中不要出现名称）。

5. 请根据上述信息，返回如下JSON格式：
{
  "分析":"请在这里填写分析", 
  "理由":"我的时间实体对应的朝代最前/偏前/居中/偏后/最后等", 
  "主题实体": "请在这里填写主题实体"
}
\end{verbatim}
}
\end{tcolorbox}
% \caption{A JSON-format case in intangible cultural heritage entity.}
\label{fig:game_step1}
\end{figure*}

\begin{figure*}[t]
\begin{tcolorbox}[title={\textbf{\small Prompt For Step2 of Timeline Ito Game}}, colback=whitesmoke, colframe=lightred, boxrule=2pt, arc=0mm]
{\small
\begin{verbatim}
基于之前的交流记录：{public_memory_str}
我的ID是：Agent {self.agent_id}，请分析我选择的实体在所有Agent中的排序。
如果在上一轮中，“本轮排序错误，错误的玩家：”后不包含我的ID，则我的排序值维持上一轮的排序值。

1. 如果主题是”水果大小“，则水果越小，排序越小；如果是“书本厚度”，则书本越薄，排序越小；如果是“船只大小”，则船只越小，排序越小；如果是“电子产品”大小，则电子产品越小，排序越小；如果是“行星大小”，则行星越小，排序越小；如果是动物大小，则动物越小，排序越小；如果是“建筑高度”，则建筑越矮，排序越小；如果是“水的温度”，则水越冷，排序越小。
本轮的主题是{theme}，返回我的“主题实体”在该主题下在所有Agent中的排序”，排序值为整数，最小为1。

2. 之前轮次我的个人记忆（这部分不公开，大家互相猜测对齐）：
{self.memory}

3. 请根据上述信息，返回如下JSON格式：
{
  "我的排序": "请在这里填写排序"
}
\end{verbatim}
}
\end{tcolorbox}
% \caption{A JSON-format case in intangible cultural heritage entity.}
\label{fig:game_step2}
\end{figure*}

\section{Timeline Ito Game Performance}\label{app:ito_acc}
The detailed performance across difficulty levels is shown in Table~\ref{tab:ito_details_performance}.
The difficulty level is determined based on the number of entities, where 3 corresponds to easy, 4 to medium, and 5 to hard.
This number also represents the number of agents.
\begin{table}[t]
\centering
\caption{ 
\textbf{Main results on Timeline Ito Game} within CTM benchmark.}
\small
\setlength\tabcolsep{2pt}
\resizebox{0.5\textwidth}{!}{%
\begin{tabular}{l|cc|cc|cc|cc}
\toprule

\multicolumn{1}{c}{\multirow{3}{*}{\textbf{Method}}} & \multicolumn{2}{c|}{\textbf{Easy}} & \multicolumn{2}{c|}{\textbf{Medium}} & \multicolumn{2}{c|}{\textbf{Hard}} & \multicolumn{2}{c}{\textbf{Overall}}\\
\cmidrule{2-9}
& Pass@3 & Pass@8 & Pass@3 & Pass@8 & Pass@3 & Pass@8 & Pass@3 & Pass@8\\
\midrule
GPT-4o & 55.00 & 80.00 & 20.00 & 30.00 & 5.00 & 10.00 & 26.67 & 40.00\\

\arrayrulecolor{black!20}\midrule
Qwen-max & 25.00 & 35.00 & 10.00 & 10.00 & 10.00 & 15.00 & 15.00 & 20.00\\

\arrayrulecolor{black}\midrule
LLaMA3.1$_{\text{8b}}$ & 0.00 & 0.00 & 0.00 & 0.00 & 0.00 & 0.00 & 0.00 & 0.00\\

\arrayrulecolor{black!20}\midrule
ChatGLM3$_{\text{6b}}$ & 5.00 & 5.00 & 0.00 & 0.00 & 0.00 & 0.00 & 1.67 & 1.67\\

\arrayrulecolor{black!20}\midrule
InternLM2.5$_{\text{7b}}$ & 5.00 & 15.00 & 0.00 & 0.00 & 0.00 & 0.00 & 1.67 & 5.00\\

\arrayrulecolor{black}\midrule
Qwen2.5$_{\text{7b}}$ & 0.00 & 15.00 & 5.00 & 5.00 & 0.00 & 0.00 & 1.67 & 6.67\\

\arrayrulecolor{black!20}\midrule
Qwen2.5$_{\text{14b}}$ & 15.00 & 20.00 & 0.00 & 0.00 & 0.00 & 0.00 & 5.00 & 6.67\\

\arrayrulecolor{black!20}\midrule
Qwen2.5$_{\text{32b}}$ & 40.00 & 50.00 & 5.00 & 15.00 & 0.00 & 0.00 & 15.00 & 21.67\\

\arrayrulecolor{black!20}\midrule
Qwen2.5$_{\text{72b}}$ & 40.00 & 55.00 & 10.00 & 10.00 & 0.00 & 5.00 & 16.67 & 23.33\\

\arrayrulecolor{black}\midrule
\bottomrule
\end{tabular}
}
\label{tab:ito_details_performance}
\end{table}

\section{Open-Book Performance}
Detailed results across tasks and entity numbers ars shown in Table~\ref{tab:openbook-qa-results}.
\label{app:openbook}

\begin{table*}[t]
\centering
\caption{ 
Detailed results under the open-book setting.}
\small
\setlength\tabcolsep{2pt}
\resizebox{\textwidth}{!}{%

\begin{tabular}{l|ccccc|ccccccc|c}
\toprule

\multicolumn{1}{c|}{\multirow{3}{*}{\textbf{Method}}} & \multicolumn{5}{c|}{\textbf{Cross Temp Count}} & \multicolumn{7}{c|}{\textbf{Question Type}} & \multicolumn{1}{c}{\multirow{3}{*}{\textbf{Avg.}}} \\
\cmidrule{2-13}
 & $=1$ (EDD) & $=2$ & $=3$ & $\geq 4$ & $\geq 4_{L}$ (LSEC) & PJ & TOU & RR & SEC & EEU & TIC & TES\\
\midrule
GPT-4o & 56.52 & 51.12 & 44.76 & 26.10 & 53.60
& 58.64 & 38.42 & 57.26 & 36.15 & 40.58 & 15.36 & 59.31 & 46.20\\
\addMethod{Openbook} & 57.76\tiny\textcolor{red}{+1.24} & 53.40\tiny\textcolor{red}{+2.28} & 45.52\tiny\textcolor{red}{+0.76} & 26.90\tiny\textcolor{red}{+0.80} & 56.80\tiny\textcolor{red}{+3.20} 
& 59.00\tiny\textcolor{red}{+0.36} & 38.72\tiny\textcolor{red}{+0.30} & 54.66\tiny\textcolor[HTML]{206546}{-2.60} & 45.30\tiny\textcolor{red}{+9.15} & 42.61\tiny\textcolor{red}{+2.03} & 17.20\tiny\textcolor{red}{+1.84} & 58.39\tiny\textcolor[HTML]{206546}{-0.92} & 49.41\tiny\textcolor{red}{+3.21} \\

\arrayrulecolor{black}\midrule
Qwen2.5$_{\text{7b}}$ & 51.80 & 39.88 & 35.96 & 12.40 & 30.00 
& 46.28 & 26.38 & 46.28 & 24.14 & 36.23 & 7.35 & 52.01 & 38.76\\
\addMethod{Openbook} & 48.64\tiny\textcolor[HTML]{206546}{-3.16} & 39.92\tiny\textcolor{red}{+0.04} & 31.88\tiny\textcolor[HTML]{206546}{-4.08} & 17.90\tiny\textcolor{red}{+5.50} & 31.60\tiny\textcolor{red}{+1.60} 
& 47.63\tiny\textcolor{red}{+1.35} & 27.89\tiny\textcolor{red}{+1.51} & 42.15\tiny\textcolor[HTML]{206546}{-4.13} & 26.04\tiny\textcolor{red}{+1.90} & 31.88\tiny\textcolor[HTML]{206546}{-4.35} & 5.84\tiny\textcolor[HTML]{206546}{-1.51} & 44.53\tiny\textcolor[HTML]{206546}{-7.48} & 37.39\tiny\textcolor[HTML]{206546}{-1.37} \\

\arrayrulecolor{black!20}\midrule
Qwen2.5$_{\text{14b}}$ & 54.36 & 51.16 & 42.56 & 23.80 & 42.00 
& 57.44 & 36.86 & 51.83 & 36.90 & 39.07 & 18.26 & 58.58 & 46.32\\
\addMethod{Openbook} & 54.32\tiny\textcolor[HTML]{206546}{-0.04} & 51.28\tiny\textcolor{red}{+0.12} & 41.76\tiny\textcolor[HTML]{206546}{-0.80} & 23.60\tiny\textcolor[HTML]{206546}{-0.20} & 44.40\tiny\textcolor{red}{+2.40} 
& 58.82\tiny\textcolor{red}{+1.38} & 36.48\tiny\textcolor[HTML]{206546}{-0.38} & 51.83\tiny\textcolor{red}{+0.00} & 39.95\tiny\textcolor{red}{+3.05} & 39.71\tiny\textcolor{red}{+0.64} & 13.86\tiny\textcolor[HTML]{206546}{-4.40} & 52.92\tiny\textcolor[HTML]{206546}{-5.66} & 46.14\tiny\textcolor[HTML]{206546}{-0.18} \\

\arrayrulecolor{black!20}\midrule
Qwen2.5$_{\text{32b}}$ & 56.28 & 52.78 & 46.24 & 26.90 & 46.40 
& 60.66 & 38.54 & 56.79 & 39.12 & 43.77 & 20.10 & 60.04 & 48.83\\
\addMethod{Openbook} & 57.92\tiny\textcolor{red}{+1.64} & 53.32\tiny\textcolor{red}{+0.54} & 46.16\tiny\textcolor[HTML]{206546}{-0.08} & 26.80\tiny\textcolor[HTML]{206546}{-0.10} & 50.80\tiny\textcolor{red}{+4.40} 
& 61.15\tiny\textcolor{red}{+0.49} & 39.93\tiny\textcolor{red}{+1.39} & 55.61\tiny\textcolor[HTML]{206546}{-1.18} & 40.67\tiny\textcolor{red}{+1.55} & 45.22\tiny\textcolor{red}{+1.45} & 16.86\tiny\textcolor[HTML]{206546}{-3.24} & 58.21\tiny\textcolor[HTML]{206546}{-1.83} & 49.51\tiny\textcolor{red}{+0.68} \\

\arrayrulecolor{black!20}\midrule
Qwen2.5$_{\text{72b}}$ & 58.20 & 48.76 & 46.84 & 31.30 & 60.80
& 61.38 & 40.77 & 54.31 & 36.62 & 42.03 & 11.52 & 62.23 & 49.30\\
\addMethod{Openbook} & 57.96\tiny\textcolor[HTML]{206546}{-0.24} & 52.00\tiny\textcolor{red}{+3.24} & 48.04\tiny\textcolor{red}{+1.20} & 30.60\tiny\textcolor[HTML]{206546}{-0.70} & 63.60\tiny\textcolor{red}{+2.80} 
& 62.67\tiny\textcolor{red}{+1.29} & 42.86\tiny\textcolor{red}{+2.09} & 54.07\tiny\textcolor[HTML]{206546}{-0.24} & 41.26\tiny\textcolor{red}{+4.64} & 44.64\tiny\textcolor{red}{+2.61} & 18.03\tiny\textcolor{red}{+6.51} & 56.75\tiny\textcolor[HTML]{206546}{-5.48} & 50.51\tiny\textcolor{red}{+1.21} \\

\arrayrulecolor{black}\midrule
\bottomrule
\end{tabular}
}
\label{tab:openbook-qa-results}
\end{table*}

\begin{figure*}[t]
    \centering
    \includegraphics[width=1\textwidth]{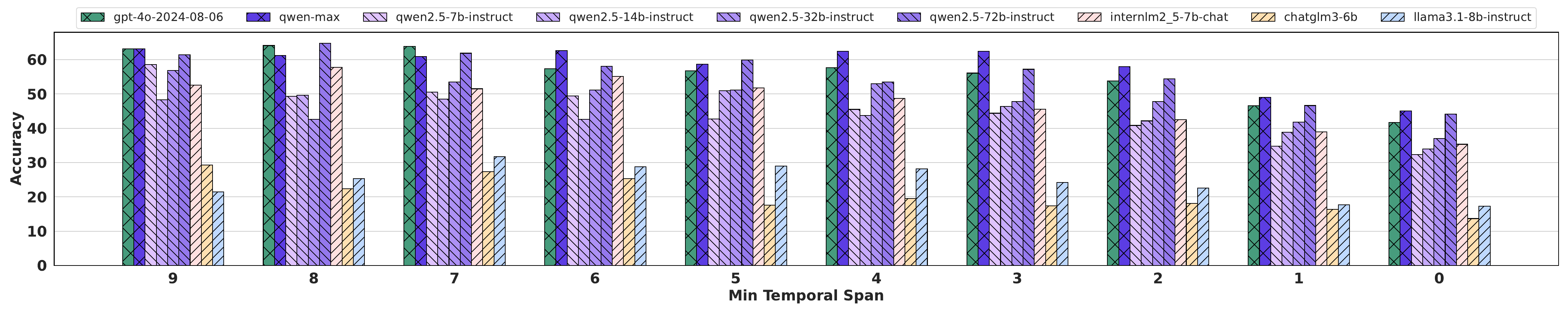}
    \caption{Accuracy across entity inter-dynastic intervals under COT prompting setting.}
    \label{fig:acc_span_cot}
\end{figure*}

\begin{figure}[ht]
    \centering
    \includegraphics[width=0.5\textwidth]{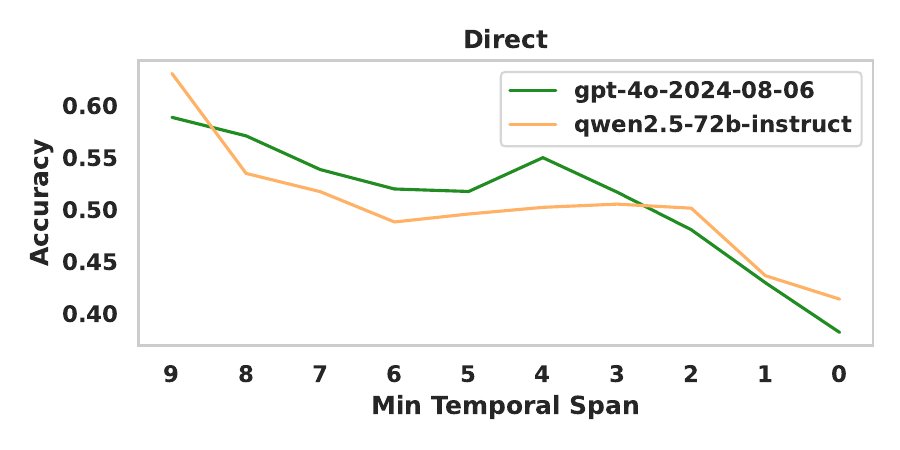}
    \caption{Accuracy across entity inter-dynastic intervals under direct prompting setting on GPT-4o and Qwen2.5-7B.
    }
    \label{fig:line_direct}
\end{figure}

\begin{figure}[ht]
    \centering
    \includegraphics[width=0.5\textwidth]{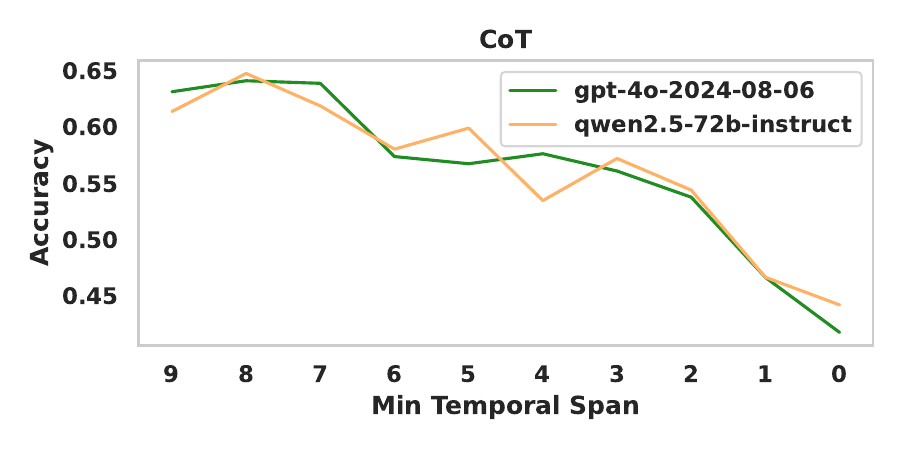}
    \caption{Accuracy across entity inter-dynastic intervals under CoT prompting setting on GPT-4o and Qwen2.5-7B.
    }
    \label{fig:line_cot}
\end{figure}

\end{CJK*}
\end{document}